\documentclass[fleqn,10pt]{wlscirep}
\usepackage[utf8]{inputenc}
\usepackage[T1]{fontenc}
\usepackage[table]{xcolor}
\title{Compact convolutional neural networks for AI-based drone detection system}

\author[1,+,*]{Gábor Farkas}
\author[1,+]{Gábor Fazekas}
\author[2,+]{Patrik Karakai}
\author[1,+]{András Németh}
\author[2,+]{Gábor Farkas}
\affil[1]{Department of Electronic Warfare, Ludovika University of Public Service, Budapest, H-1083, Hungary}
\affil[2]{Faculty of Informatics, Eötvös Loránd University, Budapest, H-1117, Hungary}

\affil[*]{farkas2.gabor@stud.uni-nke.hu}

\affil[+]{these authors contributed equally to this work}


\begin{abstract}
The increasing use of first-person-view drones in modern conflicts has created a demand for compact and reliable detection systems capable of operating in complex electromagnetic environments. These drones continuously transmit video signals through onboard video transmitters, generating radio-frequency emissions that can be exploited for early detection. This study investigates the use of lightweight convolutional neural networks for automated detection of drone signals captured by a software-defined radio-based electronic warfare framework. Samples are converted into rasterized time-domain images, providing a computationally efficient input representation suitable for embedded systems. Several custom model architectures were designed and benchmarked in terms of accuracy, model size, and inference performance using a dataset containing approximately 40,000 labeled images. In addition to offline testing, the models were integrated into a GNU Radio signal processing chain for real-time evaluation. The results show that compact models can achieve high detection accuracy while maintaining low computational requirements, making them suitable for embedded radio-frequency monitoring applications. Compared with existing spectrogram-based RF detection methods, the proposed approach eliminates frequency-domain preprocessing and achieves comparable accuracy with significantly reduced computational cost.
\end{abstract}
\begin{document}

\flushbottom
\maketitle
%
%
\thispagestyle{empty}

\section*{Introduction}
The increasing availability of unmanned aerial systems (UAS), particularly first-person-view (FPV) drones, has substantially transformed the operational conditions of contemporary battlefields\cite{ref11}. Although these platforms were originally designed for civilian and recreational applications, their affordability, accessibility, and ease of modification have led to their widespread adaptation for military purposes \cite{ref10, ref14}. Recent armed conflicts, especially the Russo–Ukrainian war, have clearly demonstrated the growing role of such systems in reconnaissance, target acquisition, and direct attacks against personnel and equipment \cite{ref12}. Due to their relatively low cost and expendable nature, FPV drones can be deployed in large numbers, enabling adversaries to create persistent and asymmetric threats against ground forces. Detecting these platforms poses several technical challenges \cite{ref13}. In contrast to larger unmanned aircraft, FPV drones typically operate at very low altitudes, have minimal radar cross-sections, and use commercially available radio-frequency (RF) communication components \cite{ref15}. These characteristics significantly reduce the effectiveness of traditional air-defense and surveillance systems, which are often optimized for larger aerial targets \cite{ref16}. For this reason, the early identification of approaching drones has become a critical factor in force protection, as it provides additional time for defensive reactions and countermeasure deployment \cite{ref17}. Within this context, electronic support measures (ESM) offer a viable detection approach by monitoring the electromagnetic spectrum and identifying RF emissions generated by drone communication links \cite{ref38}.

A characteristic feature of FPV drones is the continuous video transmission between the drone and the remote operator by an onboard video transmitter (VTX) \cite{ref24}. The emitted signals exhibit repetitive structures related to the underlying video transmission scheme, making them suitable candidates for RF-based detection techniques \cite{ref10}. In earlier stages of this research, a detection method based on autocorrelation was implemented to identify the periodic characteristics of the signals. While the approach proved capable of detecting certain signal types, practical experiments revealed several limitations. In particular, the method is sensitive to noise, signal distortion, and variations in transmission parameters, which reduces its robustness and adaptability. To overcome these limitations, alternative detection approaches must be investigated that can better tolerate signal variability and environmental interference. Artificial intelligence (AI), and in particular machine learning (ML) techniques, have recently demonstrated significant potential in RF-based drone detection and classification tasks \cite{ref18, ref19, ref20}. Initial experiments using existing, general-purpose CNN architectures demonstrated that such networks are capable of recognizing the rasterized RF patterns associated with FPV video transmissions. However, widely used models designed for large-scale image classification tasks, such as ResNet-based architectures, are typically too computationally demanding for deployment on compact, low-power embedded platforms. Their large parameter count and high memory requirements limit their suitability for portable electronic warfare systems intended for field use. The central hypothesis of this work is that the repetitive temporal structure of analog FPV video transmissions can be exploited directly in the time domain without resorting to computationally expensive transformations into the frequency domain. If successful, such an approach could enable RF-based drone detection using significantly smaller neural networks and lower preprocessing overhead than commonly reported spectrogram-based methods.

This work extends our previous studies \cite{ref10}, in which a conceptual SDR-based FPV drone detection framework and a preliminary CNN-based classification approach were introduced. In that earlier work, the focus was primarily on system-level design and feasibility demonstration, including signal acquisition and a proof-of-concept neural network implementation. In contrast, the present paper provides a systematic and quantitative investigation of lightweight CNN architectures tailored for embedded deployment. Multiple custom-designed CNN models are developed and compared, and a detailed benchmarking methodology is introduced to evaluate accuracy, computational complexity, and real-time performance within a GNU Radio environment. Furthermore, this study highlights the discrepancy between offline evaluation and live system behavior, which was not addressed previously. These contributions establish a clear advancement beyond the initial feasibility study toward a practically deployable, resource-efficient AI-based RF detection solution.

\subsection*{Related Work}
RF-based drone detection has attracted considerable attention in recent years due to the increasing use of commercial and military unmanned aerial systems. One major research direction employs deep-learning models operating on captured RF communication signals, learning discriminative representations for drone identification and classification \cite{ref37}. These approaches have demonstrated high classification performance under controlled conditions and are capable of distinguishing between different drone platforms and communication devices \cite{ref32, ref33}. Public datasets such as DroneRF have become widely used benchmarks for RF-based drone detection and identification research \cite{ref36}.

\begin{figure}[ht!]
    \centering
    \includegraphics[width=0.6\linewidth]{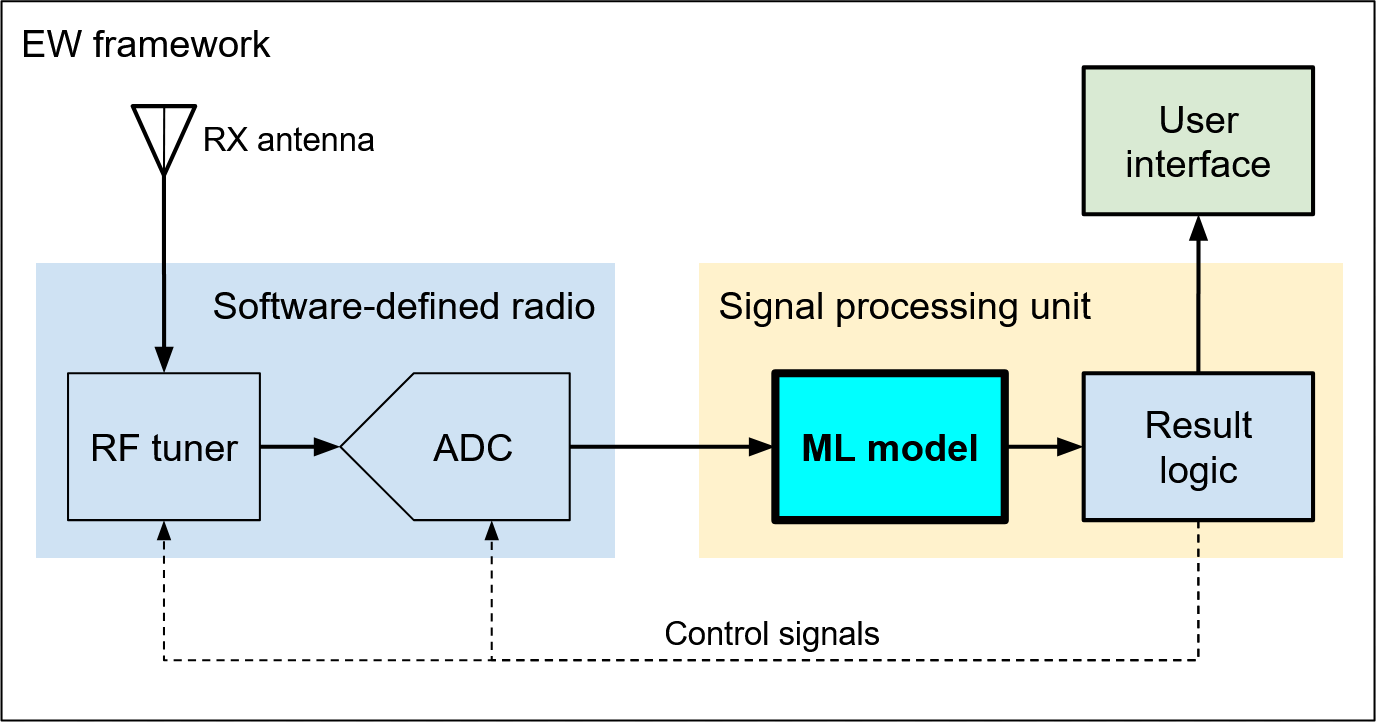}
    \caption{Simplified block diagram of the proposed electronic warfare framework. The software-defined radio (SDR) performs RF acquisition and digitization, followed by preprocessing and rasterization of the signal stream. The CNN-based classification module is highlighted, showing its role in real-time detection of FPV drone transmissions within the processing chain.}
    \label{fig:1}
\end{figure}

A second major category relies on frequency-domain representations. Spectrogram-based CNN architectures and related time-frequency methods have demonstrated classification accuracies exceeding 95\% by exploiting spectral characteristics of drone communication signals \cite{ref34}. Frequency-domain methods are generally robust against noise and interference because frequency-domain preprocessing isolates discriminative signal components before classification. However, they typically require computationally intensive preprocessing steps such as short-time Fourier transform (STFT) or other Fourier-based operations. As a result, the computational requirements of both preprocessing and classification may become problematic for resource-constrained embedded electronic warfare platforms.

More recently, several studies have investigated direct processing of RF streams using lightweight neural architectures \cite{ref33, ref34}. These methods aim to reduce preprocessing overhead while maintaining competitive classification performance. The approach proposed in this work follows a similar philosophy but employs a rasterized time-domain representation specifically tailored to analog FPV video transmissions. Instead of extracting handcrafted RF fingerprints or generating spectrograms, the received signal is converted directly into an image-like representation that preserves repetitive temporal structures. This design is motivated by the observation that analog FPV video signals exhibit highly repetitive temporal structures that can be exploited without explicit frequency-domain analysis. Consequently, the proposed method bridges the gap between computationally intensive spectrogram-based approaches and traditional RF fingerprinting techniques, providing a favorable compromise between computational efficiency and detection performance for embedded electronic warfare systems.

\section*{Environmental Requirements}
The development of an AI–based drone detection capability must be considered within the context of the electronic warfare (EW) framework in which it is intended to operate. In the proposed system architecture RF signals are captured using a SDR connected to an embedded signal processing unit. The SDR performs the initial analog processing stages, including amplification, filtering, and down-conversion, before digitizing the received signal through an analog-to-digital converter (ADC). The resulting digital RF stream is then forwarded to the processing unit, where detection algorithms are executed in real-time \cite{ref35, ref38}. Within this architecture, the machine learning (ML) model, more precisely the CNN, functions as a signal classification module that operates on preprocessed representations of the captured RF data as shown in Fig. \ref{fig:1}.

The primary role of the CNN is to determine whether the currently observed RF stream contains characteristics associated with FPV drone video transmissions. Consequently, the neural network must be integrated into the processing chain in a manner that allows continuous analysis of the incoming signal while maintaining the real-time operational requirements of the system. Since the intended platform is a compact, portable EW device, the available computational resources are limited. Therefore, both the preprocessing pipeline and the neural network architecture must be designed with computational efficiency in mind.

A key design decision concerns the representation of RF signals provided as input to the neural network. The method employed in this research operates entirely in the time domain by converting the RF signal stream directly into rasterized image representations. Rasterization involves arranging sequential samples from the RF stream into a two-dimensional grid structure that resembles an image. This representation preserves temporal patterns and repetitive signal structures while avoiding the need for computationally expensive spectral analysis. For signals with strong periodic characteristics, such as analog FPV video transmissions, the resulting rasterized images contain distinctive visual patterns that can be effectively recognized by a CNN. An additional advantage of this approach is its simplicity, as the preprocessing primarily consists of magnitude calculation and sequential sample mapping, both of which are computationally lightweight operations.

\begin{figure}[ht!]
    \centering
    \includegraphics[width=0.6\linewidth]{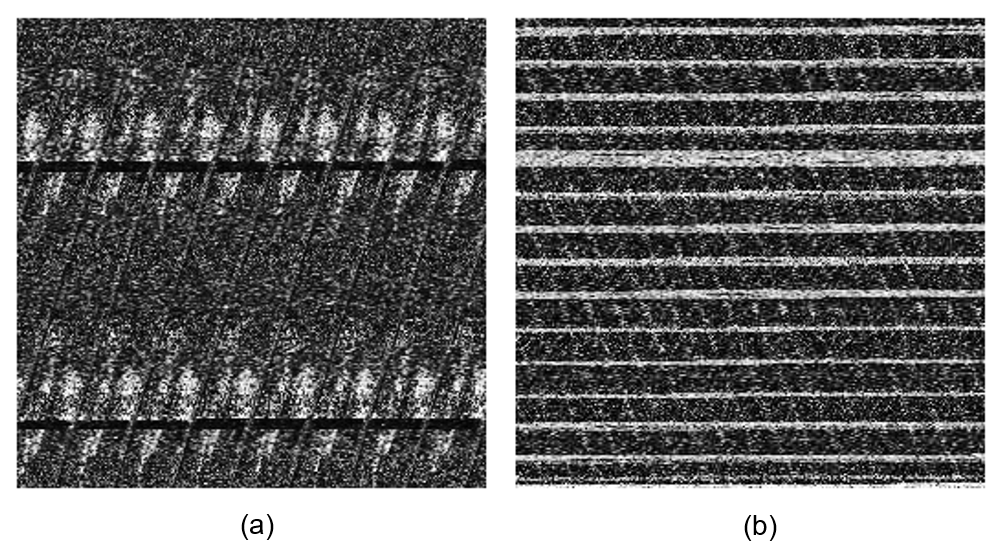}
    \caption{Rasterized representations of RF signal segments used as CNN input: (a) FPV drone video transmitter signal exhibiting regular horizontal and vertical structures caused by the underlying video transmission scheme. (b) Non-target RF signal lacking such periodic patterns. The visual distinction between structured and unstructured patterns forms the basis for CNN-based classification.}
    \label{fig:2}
\end{figure}

Beyond computational efficiency, rasterized representations provide practical advantages during dataset generation and validation. Since the RF stream is converted into image-like structures, individual samples can be visually inspected, enabling manual verification of signal characteristics and labeling accuracy. This visual interpretability facilitates better control over the dataset creation process and helps ensure that the training data accurately reflects the signal patterns of interest. Such control is particularly valuable in RF classification problems, where publicly available labeled datasets are often limited or entirely unavailable.
To standardize the input format for neural network training and evaluation, a fixed image resolution is used. In the proposed implementation, each rasterized RF sample is represented as a grayscale image with a resolution of 256$\times$256 pixels as shown in Fig. \ref{fig:2}. This size provides a sufficient balance between preserving signal structure and maintaining manageable computational complexity for embedded processing. Each image therefore represents a short segment of the captured RF stream, transformed into a format suitable for CNN-based pattern recognition.

The output of the neural network is defined as a probability value indicating the likelihood that the analyzed RF segment corresponds to an FPV drone video transmission. In practical terms, the model performs a binary classification task in which the input image is categorized either as containing the target signal or as representing other RF activity. The network output is therefore typically expressed as a probability score, which can then be evaluated against a predefined decision threshold to determine whether a detection event should be triggered.
Within the overall EW framework, this probability output is forwarded to a higher-level decision logic module responsible for generating system responses. Depending on the operational configuration, these responses may include visual or acoustic alerts, logging of detection events, or the activation of additional countermeasure systems. The modular design of the framework ensures that the CNN-based classifier can be integrated without significantly altering the surrounding signal processing architecture.

\begin{figure}[ht!]
    \centering
    \includegraphics[width=0.7\linewidth]{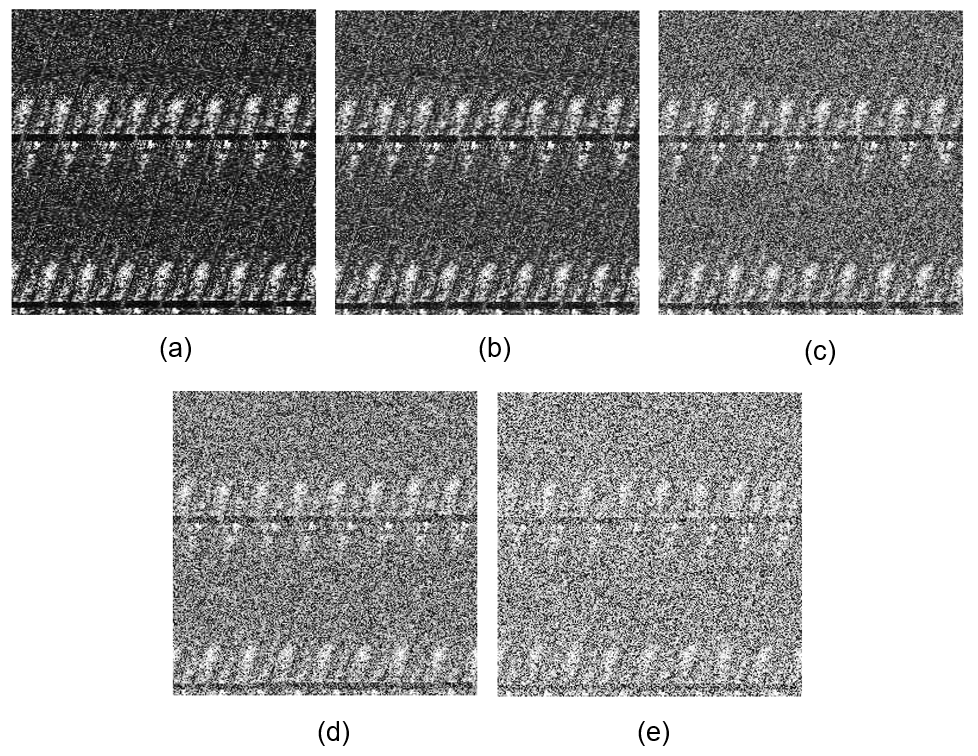}
    \caption{Examples of synthetic noise augmentation applied during dataset generation: (a) 10\% noise, (b) 20\% noise, (c) 30\% noise, (d) 40\% noise, (e) 50\% noise}
    \label{fig:3}
\end{figure}

\section*{Dataset generation}
The effectiveness of ML–based signal detection methods largely depends on the quality and representativeness of the training dataset. However, in the domain of RF-based drone detection, publicly available labeled datasets are extremely limited. As a result, the dataset used in this study was generated using RF recordings obtained within the previously developed electronic warfare framework.
The dataset was created using a SDR front-end capturing real FPV drone VTX signals. The positive-class dataset originated from an approximately 20-second recording captured from a BetaFPV A03 5.8 GHz analog video transmitter. A separate RF recording of similar duration, containing background activity without FPV video transmissions was used to generate the negative class. The raw RF streams were first converted into rasterized image representations as described in the previous section. Each rasterized image represents a short temporal segment of the RF signal arranged into a fixed two-dimensional structure.

The original recordings were divided into 100 non-overlapping base chunks before any augmentation was performed. These chunks were assigned to the training, validation, and test partitions using an 80/10/10 split. All augmentation procedures were subsequently applied independently within each partition. Consequently, augmented samples derived from a given base chunk never appeared in multiple subsets.

The use of a relatively short RF capture reflects the intended operational use case of rapidly constructing a usable detector from limited captured RF data. The live evaluation experiments presented later in this paper were conducted using a different FPV platform (BetaFPV Meteor75 Pro) than the BetaFPV A03 analog video transmitter used for training data generation, providing an initial assessment of cross-platform generalization.

Augmentation techniques were applied to increase dataset diversity and to simulate the variability encountered in real-world electromagnetic environments \cite{ref22}. The augmentation parameters were selected based on expected variations in RF signal conditions, such as changes in signal strength, noise levels, and temporal alignment within the sampled stream. These transformations enable the training process to expose the neural networks to a broader range of signal conditions than would be possible using the original recordings alone. The augmentation methods included controlled temporal shifting of the signal segments, resampling variations, and the addition of synthetic noise to approximate realistic reception conditions as shown in Fig. \ref{fig:3}. The augmentation parameters used during dataset generation are summarized in Tab. \ref{tab:5}. Resampling within a $\pm$0.5\% range was applied to emulate small timing variations of analog FPV video transmitters, while five noise levels were used to represent a range of signal-to-noise conditions.

\begin{table}[ht!]
\caption{Signal-domain augmentation parameters used during dataset generation}
\label{tab:5}
\centering
\begin{tabular}{|p{100pt}|p{100pt}|}
\hline
\rowcolor{lightgray}
Parameter & Value \\
\hline
Image resolution & 256 $\times$ 256 \\
\hline
Samples per image & 65,536 \\
\hline
Sampling rate & 1 MSPS \\
\hline
Resampling range & $\pm$0.5\% \\
\hline
Resampling steps & 100 \\
\hline
Temporal offset & Random chunk position \\
\hline
Noise levels & 5 \\
\hline
Relative noise distortion & 10--50\% \\
\hline
\end{tabular}
\end{table}

Through the application of these augmentation techniques, a dataset containing approximately 40,000 rasterized images was generated. The dataset was organized into two primary classes. The first class, labeled \textit{signal}, contains samples representing RF segments in which FPV drone video transmissions are present. The second class, labeled \textit{other}, contains samples representing background RF activity or signal segments where the target transmission is absent. This binary classification structure reflects the operational requirement of the detection system, where the primary objective is to determine whether a monitored RF channel contains a drone-related transmission. 

As described previously, the base chunks were partitioned into training, validation, and test subsets prior to augmentation. The training set contains 80\% of the total samples and is used for optimizing the parameters of the neural network models. The validation set, comprising 10\% of the dataset, is used during the training process to monitor model performance and prevent overfitting by providing an independent measure of generalization capability. The remaining 10\% of the samples form the test set, which is reserved exclusively for the final evaluation of the trained models. This separation ensures that performance metrics reported in the benchmarking stage reflect the behavior of the models on previously unseen data. The dataset used in this article for training the models is available on the GitHub repository.\cite{ref40}

\section*{Custom CNN Architectures}
To identify CNN architectures that balance detection accuracy with computational efficiency for embedded deployment, a series of custom models were designed and evaluated. The primary goal was to explore how network depth and width influence performance when classifying rasterized RF images of FPV drone video transmissions. In addition, the experiments aimed to determine architectures that minimize memory usage and processing requirements while maintaining acceptable classification accuracy.

\subsection*{Baseline Model}
\begin{figure}[ht!]
    \centering
    \includegraphics[width=0.35\linewidth]{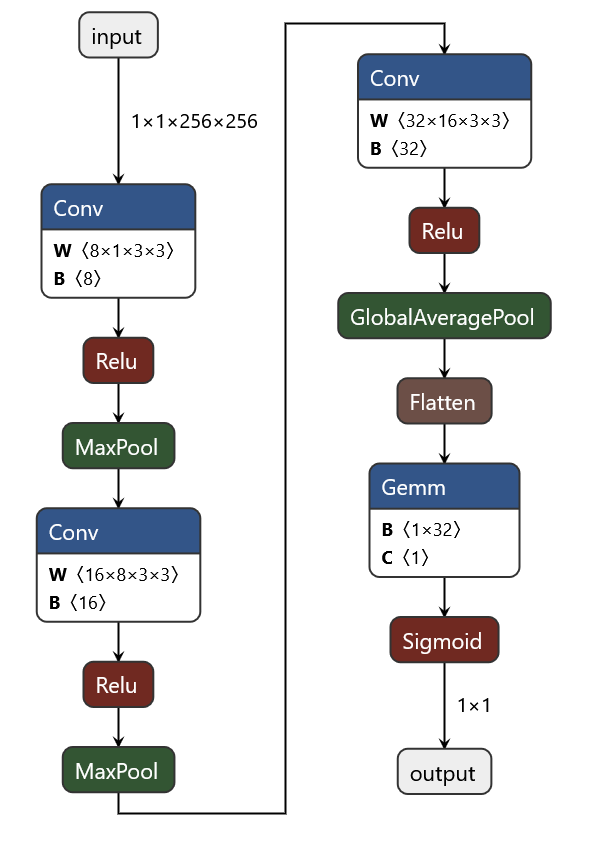}
    \caption{Structure of baseline model accepting a 256$\times$256 pixel grayscale image as input. The output is a single value representing the probability of presence of FPV drone VTX signal.}
    \label{fig:4}
\end{figure}

As a starting point, a simple three-layer CNN was implemented as a baseline model. This network consisted of three convolutional layers with channel sizes 8, 16, 32 and a pooling operation applied after every convolutional layer as shown in Fig. \ref{fig:4}. The purpose of the baseline model was to provide a reference point for both accuracy and resource utilization, against which subsequent modifications could be compared.

\begin{figure}[ht!]
    \centering
    \includegraphics[width=0.6\linewidth]{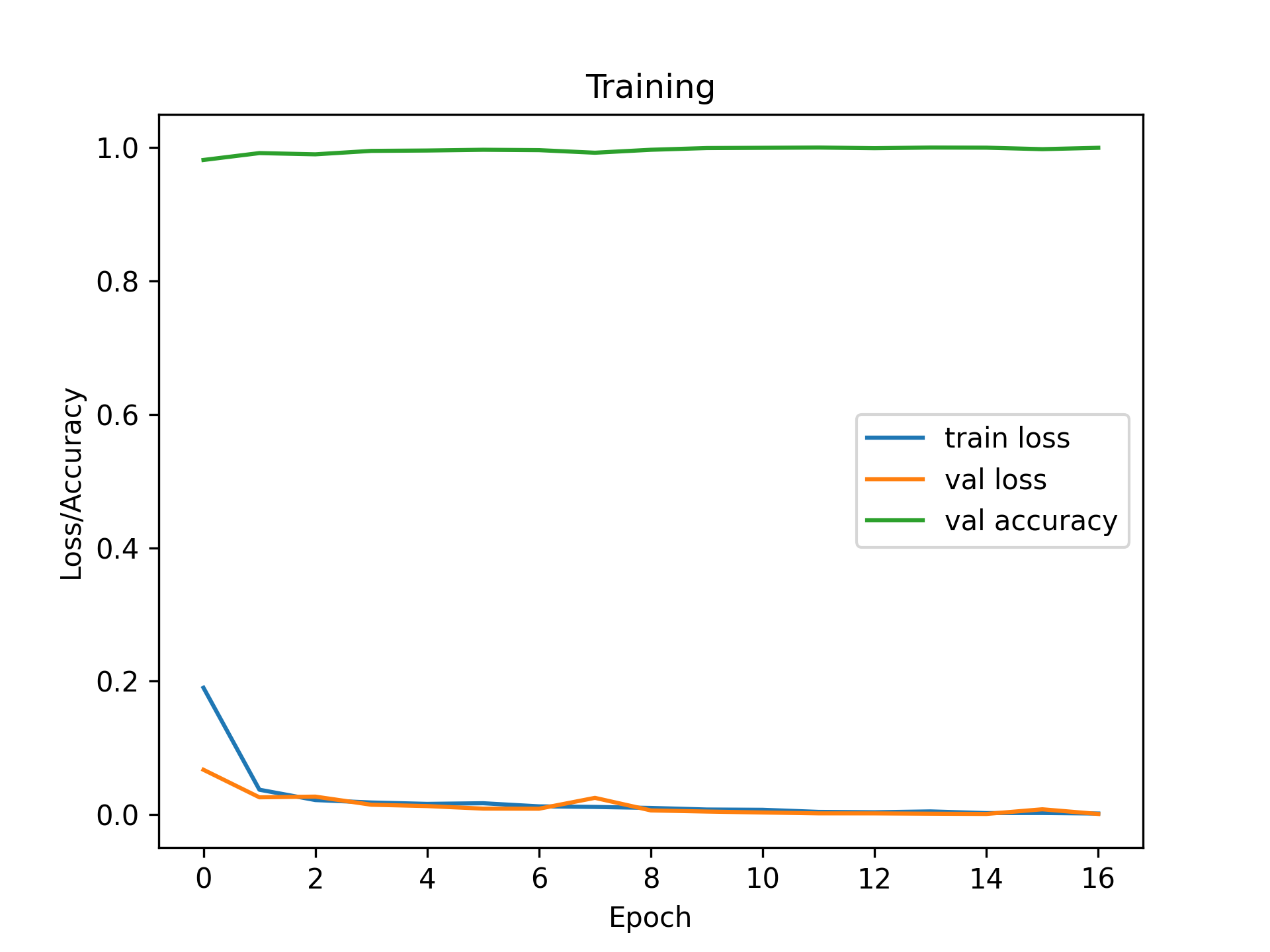}
    \caption{Training and validation accuracy of the baseline CNN model over epochs. The model rapidly converges to near-perfect accuracy, indicating strong separability of the dataset. However, as later demonstrated in live testing, high offline accuracy does not necessarily translate to reliable real-time performance.}
    \label{fig:5}
\end{figure}

Training this baseline model on the generated RF raster dataset demonstrated that a relatively shallow network is capable of distinguishing between signal and other classes. With a parameter count of 5921 and model size 25 kB, it reached 99.28\% training accuracy after 12 epochs as shown in Fig. \ref{fig:5}. Finally, the model was embedded in a GNU Radio processing stream to verify real-life functionality. Empirical tests showed that the predictions are not reliable.

\subsection*{Architectural Variations}
Based on the baseline model, multiple variations of the CNN architecture were designed to explore the trade-offs between depth and width, but keeping pooling strategy the same. Below are some considerations for constructing additional models:

\begin{itemize}
    \item Shallower networks: Reduced number of layers and channels. This configuration was intended to minimize computational cost but risked insufficient capacity to capture complex signal patterns.
    \item Wider networks: Increased the number of channels per layer while keeping a similar depth. Wider networks can encode richer feature representations but require more memory and compute.
    \item Deeper networks: Added additional convolutional layers. Deep networks provide more hierarchical feature extraction, potentially improving accuracy on more subtle RF patterns, but at the cost of higher computational demands.
    \item Medium-depth configurations: A compromise between depth and width was also tested to explore balanced performance suitable for embedded platforms.
    \item Deep-efficient models: Architectures were created to study whether multiple small convolutional layers could achieve comparable feature extraction to wider layers while limiting computational cost.
\end{itemize}

Based on these considerations, a set of models was constructed to evaluate the effects caused by the structure of CNN. The models are summarized in Tab. \ref{tab:1}.

\begin{table}[ht!]
    \centering
    \caption{Overview of the designed CNN architectures evaluated in this study. The models differ in depth (number of layers) and width (number of channels per layer), enabling systematic investigation of the trade-off between representational capacity and computational efficiency for embedded RF signal classification.}
    \label{tab:1}
    \setlength{\tabcolsep}{3pt}
    \begin{tabular}{|p{80pt}|p{75pt}|p{90pt}|p{120pt}|}
        \hline
        \rowcolor{lightgray}
            Model & Number of layers & Channel size per layer & Note\\
        \hline
            Baseline & 3 & 8-16-32 & Reference model\\ 
            \hline
            Shallow & 3 & 4-8-8 & Minimal parameters\\
            \hline
            Shallow & 3 & 8-16-16 & Slightly smaller than baseline\\
            \hline
            Wide & 3 & 16-32-64 & Larger feature maps\\
            \hline
            Deep & 6 & 8-8-16-16-16-16 & Hierarchical features\\
            \hline
            Deep-wide & 6	 & 8-16-16-32-32-32 & Combines depth and width\\
            \hline
            Medium-deep & 6 & 8-16-32-32-64-64 & Balanced performance\\
            \hline
            Deep-efficient & 8 & 8-8-8-8-16-16-16-16 & Multiple small layers\\
        \hline
    \end{tabular}
\end{table}

The selection of layers and channels reflects a trade-off between accuracy and computational efficiency. Shallow or narrow networks minimize memory footprint but may lack the capacity to distinguish subtle variations in FPV video signals. In contrast, deep or wide networks improve representation power but may increase latency and power consumption critical factors for embedded EW deployment. For our different models only architectural changes were performed regarding channel width and layer count. The complete layer configuration of each architecture is summarized in Tab. \ref{tab:1} using channel-width notation. An accompanying GitHub repository provides the implementation details required for reproducibility. For training the models we used the following hyperparameters:

\begin{itemize}
    \item Batch size: 32
    \item Maximum epochs: 100
    \item Learning rate: 0.001
\end{itemize}

\section*{Benchmarking Methodology}

To evaluate the performance of the custom CNN architectures in a realistic deployment scenario, all models were benchmarked on identical hardware using a standardized testing procedure created in Python and using ONNX Runtime. The objective of this benchmarking process was to assess classification accuracy, computational efficiency, memory usage, and real-time performance of each network. The hardware environment remained constant across all experiments to ensure fair comparison between models. During testing, the SDR-based RF capture chain provided input signals to the neural network, and the detection of the FPV drone VTX was validated by alternately enabling and disabling the transmitter. Correct system behavior was confirmed when model output probabilities reflected the presence or absence of the FPV drone VTX signal.

\begin{figure}[ht!]
    \centering\includegraphics[width=0.5\linewidth]{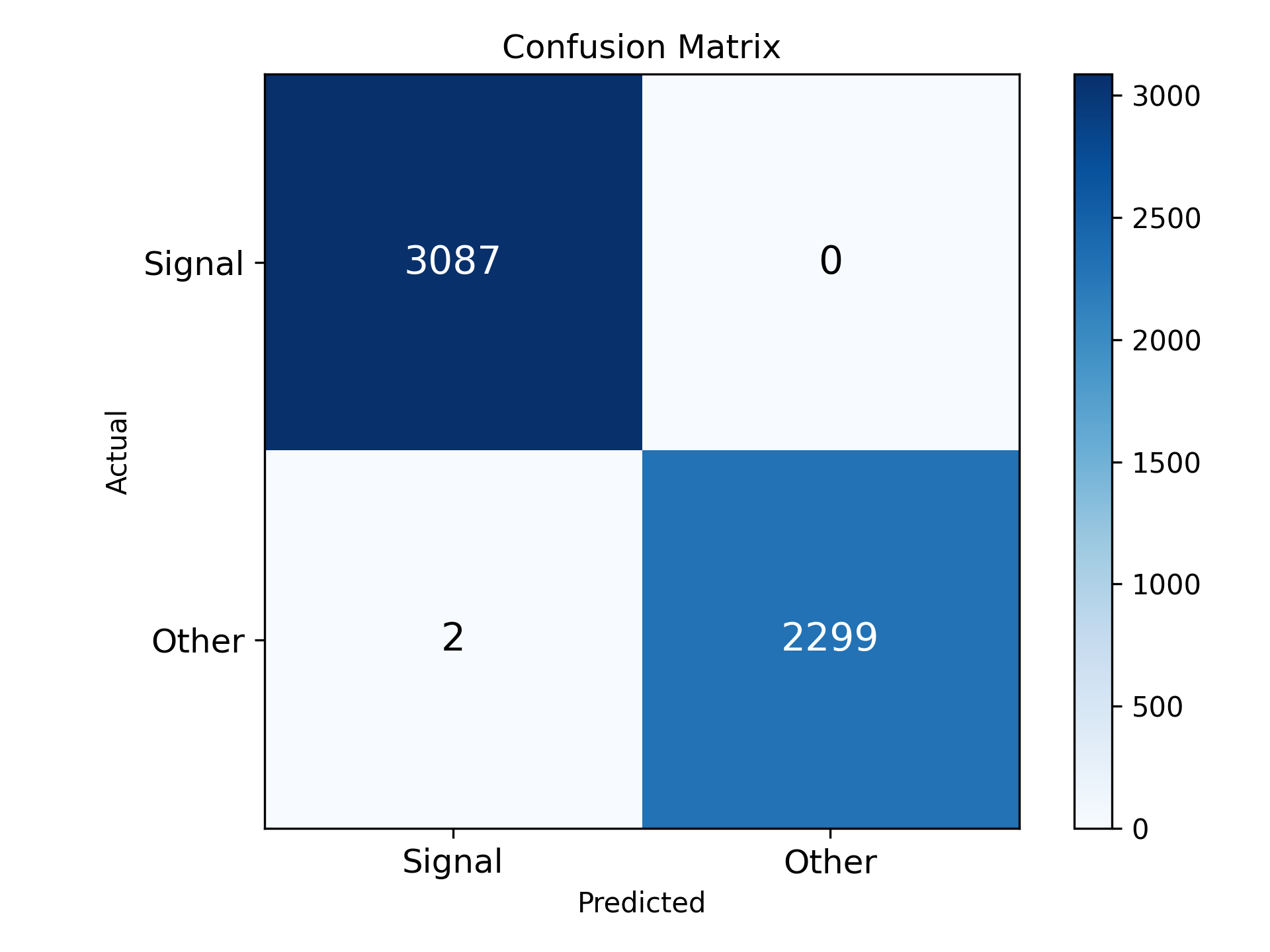}
    \caption{Confusion matrix of the baseline CNN evaluated on the test dataset. The near-perfect classification results indicate high separability between classes under offline conditions, with minimal false positives and false negatives.}
    \label{fig:6}
\end{figure}

All CNN models were exported to the ONNX format to ensure consistent inference across platforms. The testing workflow included the following steps:

\begin{table}[ht]
    \caption{Benchmark results of the evaluated CNN architectures on the offline test dataset. Metrics include model size, parameter count, inference throughput, and classification performance.}
    \centering
    \label{tab:2}
    \setlength{\tabcolsep}{3pt}
    \begin{tabular}{|p{45pt}|p{27pt}|p{35pt}|p{36pt}|p{36pt}|p{36pt}|p{35pt}|p{35pt}|p{35pt}|p{27pt}|p{30pt}|p{38pt}|}
        \hline
        \rowcolor{lightgray}
            Model & Size [MB] & Parame-ters & Through-put [img/sec] &  Accuracy (average) & Accuracy \newline (std. dev.) & Precision \newline (average) & Recall \newline (average) & F1-score \newline (average) & ROC-AUC & Epoch \newline (average) & Threshold \newline (Youden J) \\
        \hline
            C2222       & 0.005	& 489    & 2361  & 0.9905 & 0.00943 & 0.9991& 0.9335 & 0.9646 & 1      & 14.6& 0.7108\\
            \hline
            C233        & 0.006 & 929    & 2667  & 0.9910 & 0.00578 & 0.9824& 0.9902 & 0.9859 & 1      & 15.6& 0.5976\\
            \hline
            C2233       & 0.007 & 1077   & 2318  & 0.9938 & 0.00414 & 0.9894& 0.9667 & 0.9768 & 1      & 12  & 0.6075\\
            \hline
            C2334       & 0.011 & 2105   & 2637  & 0.9982 & 0.00141 & 0.9995& 0.9908 & 0.9951 & 1      & 16.2& 0.9605\\
            \hline
            C344        & 0.016 & 3585   & 2996  & 0.9943 & 0.00251 & 0.976 & 0.9918 & 0.9835 & 1      & 13.4& 0.6446\\
            \hline
            C3344       & 0.019 & 4169   & 3708  & 0.9992 & 0.00026 & 0.9989& 0.9985 & 0.9987 & 1      & 11.6& 0.6446\\
            \hline
            C345        & 0.025 & 5921   & 2605  & 0.9928 & 0.00384 & 0.9815& 0.9910 & 0.9860 & 0.9999 & 14.2 & 0.6896\\
            \hline
            C2345       & 0.026 & 6177   & 2466  & 0.9989 & 0.00110 & 0.9984& 0.9967 & 0.9975 & 1      & 15.6 & 0.9882\\
            \hline
            C3445       & 0.034 & 8241   & 2824  & 0.9993 & 0.00082 & 0.9989& 0.9948 & 0.9963 & 1      & 14.6 & 0.7719\\
            \hline
            C334444     & 0.038 & 8809   & 3293  & 0.9998 & 0.00021 & 0.9998& 0.9991 & 0.9994 & 1      & 13  & 0.9702\\
            \hline
            C33334444   & 0.043 & 9977   & 3468  & 0.9997 & 0.00052 & 0.9999& 0.9986 & 0.9992 & 1      & 11.2& 0.6705\\
            \hline
            C4455       & 0.065 & 16401  & 1704  & 0.9989 & 0.00081 & 0.9973& 0.9938 & 0.9955 & 1      & 11.2& 0.7764\\
            \hline
            C456        & 0.091 & 23361  & 911	 & 0.9968 & 0.00116 & 0.9955& 0.9903 & 0.9929 & 1      & 9.8 & 0.8363\\
            \hline
            C3456       & 0.096 & 24449  & 2053  & 0.9992 & 0.00055 & 1.0   & 0.992  & 0.9960 & 1      & 12.4  & 0.9137\\
            \hline
            C344555     & 0.106 & 26737  & 2565	 & 0.9998 & 0.00021	& 1.0   & 0.9982 & 0.9991 & 1      & 8.2  & 0.9764\\
            \hline
            C345566     & 0.273 & 70625  & 2039  & 0.9997 & 0.00009 & 0.9998& 0.9981 & 0.9989 & 1      & 26  & 0.558\\
            \hline
            LeNet5      & 0.235 & 60941  & 16579 & 0.9947 & 0.00167 & 0.9937& 0.9938 & 0.9937 & 0.9999 & 26 & 0.4329\\
            \hline
            MobileNet V2 & 8.449 & 2207521& 830   & 1.0    & 0.0     & 1.0   & 0.9999 & 0.9999 & 0.9999 & 2.2  & 0.98\\
        \hline
    \end{tabular}
\end{table}

\begin{itemize}
    \item Dataset Loading: Rasterized RF images from the test set were loaded and normalized to a range of 0-1. Resize of images was not needed because the 256$\times$256 pixel format meets the input shape of the models saving computational overhead. Labels were mapped to a binary classification scheme, where \textit{signal} represents the FPV signal and \textit{other} is for background/noise.
    \item Model Warmup: To avoid initial execution overhead skewing performance metrics, each model underwent a warmup phase consisting of repeated inference on a single input sample.
    \item Inference Measurement: For each test image, inference time was recorded to determine average latency and throughput. The total runtime across all samples was also measured.
    \item Resource Monitoring: Peak memory usage during inference was monitored using system utilities to assess suitability for embedded deployment.
    \item Accuracy Evaluation: Predictions from the CNN were compared to ground-truth labels to compute classification metrics, including accuracy, precision, recall, and confusion matrices as shown in Fig. \ref{fig:6}.
    \item Model Information Logging: Model parameters, size, and structure were recorded. Additionally, system configuration and execution environment details were saved to provide a reproducible benchmark reference.
\end{itemize}

Each CNN processed a rasterized image representation of the RF signal and produced a single probability value between 0 and 1, indicating the likelihood that an FPV video signal was present in the captured RF segment. For the offline benchmark presented in Tab. \ref{tab:2}, binary predictions were obtained using the conventional decision threshold of 0.5, enabling consistent comparison across all evaluated architectures. In addition, an optimal decision threshold was determined on the validation subset using Youden’s J statistic and is reported in Tab. \ref{tab:2}. These Youden-derived thresholds were subsequently used during the live GNU Radio evaluation presented in Tab. \ref{tab:3}. This standardized input-output interface enables straightforward integration of different CNN architectures into the SDR processing chain while facilitating direct comparison of model performance under operationally relevant conditions.

\section*{Results and Evaluation}

Benchmarking results obtained from the evaluated CNN architectures are summarized in this chapter. The objective of the evaluation was not only to compare classification accuracy but also to investigate how model size, architecture, and computational complexity influence real-world detection performance when deployed within the GNU Radio-based signal processing environment.

All models were first evaluated using the previously described benchmarking framework on the test dataset consisting of rasterized images. Tab. \ref{tab:2} summarizes the most relevant metrics, including model size, parameter count, inference throughput, classification accuracy, and other metrics. Since all accuracies were almost identical, we performed 5 training runs for each model to determine the standard deviation. A threshold analysis was also performed to determine practical values that we used during live testing. Model names are constructed from the number of layers and the width represented in power of two. For example, C233 refers to a 3 layer setup, each with a width of 4-8-8. We have selected other small size CNNs to compare their performance. Besides our own models we trained and benchmarked a LeNet5 and a MobileNet V2 model too.

The results show that all models achieved nearly perfect classification performance, typically between 99.31\% and 99.98\%. In terms of computational efficiency, a minor relationship can be observed between model complexity and inference throughput. Models with smaller width parameters generally achieve higher throughput. For instance model C3344 and C334444 achieved inference speeds exceeding 3000 images per second. In contrast, wider architectures such as C456 achieved a throughput of approximately 911 images per second. These results confirm the expected trade-off between model width and computational performance. LeNet5 achieved extremely high throughput compared to our models. However, it should be noted that the model architecture is significantly different and the input size is much smaller. MobileNet V2 gave the smallest throughput and the highest parameter count.

\begin{table}[ht!]
    \centering
    \caption{Performance of selected CNN models evaluated in a live GNU Radio RF processing environment. The results reveal that models with near-perfect offline accuracy can exhibit significantly different real-time behavior, including false positives or missed detections, demonstrating the importance of evaluating models under continuous streaming conditions.}
    \label{tab:3}
    \setlength{\tabcolsep}{3pt}
    \begin{tabular}{|p{55pt}|p{40pt}|p{37pt}|p{37pt}|p{37pt}|p{37pt}|p{40pt}|p{30pt}|p{40pt}|p{40pt}|p{35pt}|}
        \hline
        \rowcolor{lightgray}
            Model name & Threshold & True \newline positive & False \newline positive & True \newline negative & False \newline negative & Accuracy & Recall & F1-score & ROC-AUC & Latency [ms] \\
        \hline
            C233            & 0.5976 & 0   & 0   & 500 & 500 & 0.5   & 0     & 0      & 0.5    & 0.5484\\
            C344            & 0.6446 & 0   & 0   & 500 & 500 & 0.5   & 0     & 0      & 0.5    & 0.4697\\
            C345            & 0.6896 & 0   & 0   & 500 & 500 & 0.5   & 0     & 0      & 0.5    & 0.5183\\
            C456            & 0.8363 & 0   & 0   & 500 & 500 & 0.5   & 0     & 0      & 0.5    & 1.4092\\
            C334444         & 0.9702 & 500 & 491 & 9   & 0   & 0.509 & 1     & 0.6707 & 0.998  & 0.4525\\
            C344555         & 0.9764 & 500 & 0   & 500 & 0   & 1     & 1     & 1      & 1      & 0.5457\\
            C345566         & 0.558  & 428 & 0   & 500 & 72  & 0.928 & 0.856 & 0.9224 & 0.941  & 0.6428\\
            C33334444       & 0.6705 & 500 & 500 & 0   & 0   & 0.5   & 1     & 0.6667 & 0.5    & 0.4479\\
            C2222           & 0.7108 & 500 & 500 & 0   & 0   & 0.5   & 1     & 0.6667 & 1      & 0.6106\\
            C2233           & 0.6075 & 500 & 0   & 500 & 0   & 1     & 1     & 1      & 1      & 0.597\\
            C2345           & 0.9882 & 500 & 500 & 0   & 0   & 0.5   & 1     & 0.6667 & 0.5    & 0.5904\\
            C2334           & 0.9605 & 500 & 279 & 221 & 0   & 0.721 & 1     & 0.7819 & 1      & 0.5622\\
            C3445           & 0.7719 & 476 & 0   & 500 & 24  & 0.976 & 0.952 & 0.9754 & 0.988  & 0.5158\\
            C3456           & 0.9137 & 0   & 0   & 500 & 500 & 0.5   & 0     & 0      & 0.5    & 0.6206\\
            C3344           & 0.6446 & 0   & 0   & 500 & 500 & 0.5   & 0     & 0      & 0.5    & 0.4372\\
            C4455           & 0.7764 & 500 & 0   & 500 & 0   & 1     & 1     & 1      & 1      & 0.9331\\
            LeNet5          & 0.4329 & 308 & 1   & 499 & 192 & 0.807 & 0.616 & 0.7614 & 0.9976 & 0.1579\\
            MobileNetV2     & 0.98   & 104 & 64  & 436 & 396 & 0.54  & 0.208 & 0.3114 & 0.629  & 1.3883\\
            Auto-correlation& -      & 500 & 0   & 500 & 0   & 1     & 1     & 1      & 1      & 11.8191\\
        \hline
    \end{tabular}
\end{table}

While offline accuracy metrics provide valuable insight into model performance, the ultimate objective of this research is real-time detection of FPV drone signals within a live RF processing chain. Therefore, all trained models were additionally integrated into a GNU Radio signal processing pipeline and tested under live conditions. We used an Ettus USRP B210 SDR for receiving signals with the following settings:

\begin{itemize}
    \item Center frequency: 5705 MHz (FPV channel E1)
    \item Sample rate: 1 MSps
    \item Automatic gain control: Enabled 
\end{itemize}

\begin{figure}[ht!]
    \centering
    \includegraphics[width=1.0\linewidth]{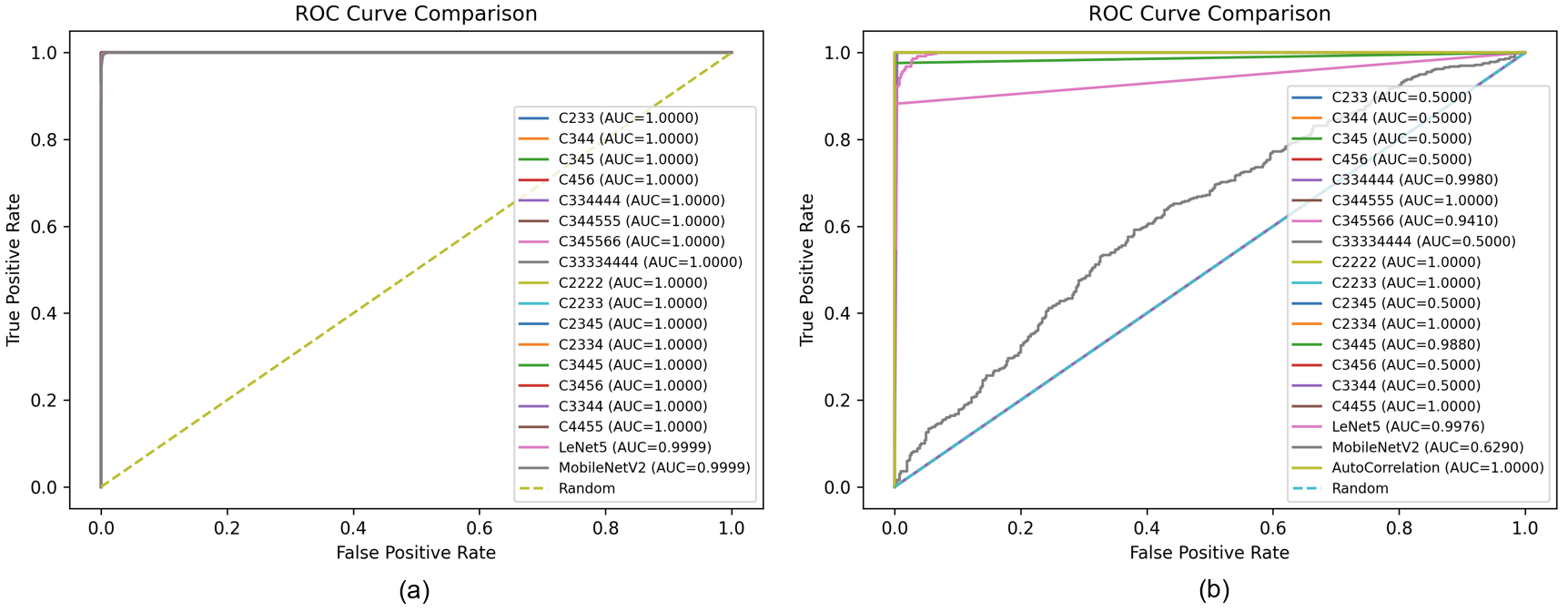}
    \caption{Comparison of ROC curves for representative CNN models under (a) offline test conditions and (b) live GNU Radio evaluation. While most models achieve near-ideal performance offline, significant degradation and variability are observed in real-time operation, highlighting the importance of evaluating RF detection models in continuous streaming environments.}
    \label{fig:7}
\end{figure}

The signal processing pipeline demodulates the incoming In-phase and Quadrature (IQ) data to magnitude, then grabs 65,536 samples from the incoming stream and performs rasterization to a 256$\times$256 shape. This is then fed into the model for processing. Using the previously determined threshold and the model output, the pipeline finally gives a true or false value as a detection result. Additionally the raw prediction values are stored for constructing the final test report. The system continuously processes the captured RF stream while the FPV drone VTX was manually switched on and off. We used a BetaFPV Meteor75 Pro (Tiny Whoop) as source of RF signal. This platform was not used during dataset generation, which exclusively relied on recordings captured from a BetaFPV A03 analog video transmitter. The live evaluation therefore extends the experimental validation to an FPV hardware configuration not represented in the training dataset. The placement of the drone and the SDR was kept consistent throughout the entire test. A correctly functioning model should produce a clear probability response corresponding to the presence or absence of the drone signal. The test results are summarized in Tab. \ref{tab:3}.
The results revealed an important observation. High offline classification accuracy does not necessarily guarantee reliable real-time behavior. Several models that achieved near-perfect accuracy during offline testing exhibited undesirable behavior in the live environment. These issues primarily manifested in two forms:

\begin{itemize}
    \item Over-sensitivity: Some models produced frequent false detections even when the FPV drone VTX signal was absent.
    \item Insufficient sensitivity: In some cases, the model failed to respond clearly when the FPV drone VTX signal was present.
\end{itemize}

\begin{figure}[ht!]
    \centering
    \includegraphics[width=0.6\linewidth]{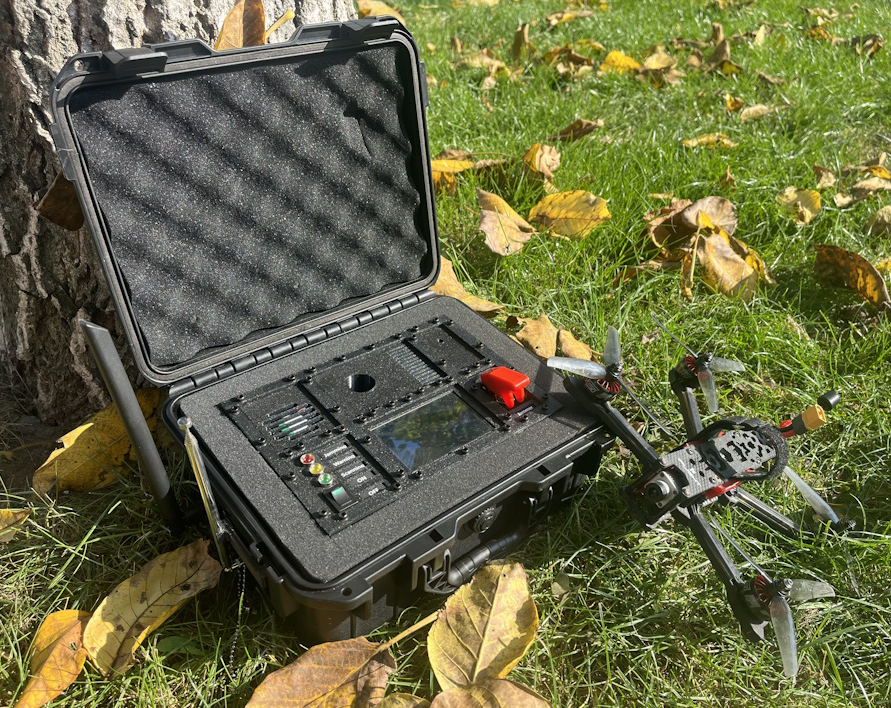}
    \caption{Portable implementation of the proposed EW framework in a rugged enclosure. The system integrates SDR hardware and an embedded processing unit capable of running the CNN-based detection pipeline in real-time, demonstrating practical deployability in field conditions.}
    \label{fig:8}
\end{figure}

For example, the model C233, despite achieving high accuracy and extremely high throughput, proved to be insufficiently sensitive during live testing, generating undesirable detection outputs. Similarly, larger architectures such as C334444 and C33334444 also demonstrated excessive sensitivity, producing frequent false positives in the GNU Radio environment. Conversely, several models demonstrated stable and reliable behavior during live operation. Architectures such as C344555, C345566, C2233, C3445 and C4455 consistently responded to the presence of the FPV drone VTX signal while maintaining low false-positive rates when the signal was absent. Interestingly, some relatively small models performed particularly well in real-time tests. For instance, C2233, which contains fewer than 1100 parameters, exhibited stable behavior and high throughput, making them especially attractive candidates for embedded implementation. The auto-correlation based detection algorithm achieved competitive detection performance under the specific conditions of the current experiment, although its latency was substantially higher than that of any CNN model tested. Fig. \ref{fig:7} provides a better visual comparison between the offline benchmark and the live test results. As a final step model C2233 was embedded in a portable version of EW framework seen in Fig. \ref{fig:8}. Since the RF hardware and the processing channel is identical to the one used for benchmarking, consistent functional results were observed.

The experiments indicate that network architecture influences not only accuracy but also the dynamic behavior of the model when operating on continuous RF streams. Extremely small networks may lack sufficient representational capacity, resulting in unstable outputs, while excessively large networks may become overly sensitive to minor signal variations. Architectures that balance moderate depth and channel width appear to provide the best compromise. Models such as C2233 and C3445 combine relatively low parameter counts with high throughput and stable real-time performance. The results demonstrate that optimizing offline accuracy alone is insufficient when designing AI-based RF detection systems. Real-time evaluation within the operational signal processing environment is essential to ensure that models behave reliably under continuous streaming conditions. 

\subsection*{Generalization and Robustness Limitations}
A further limitation of the present study is that the training dataset originated from a single FPV video transmission recording and a single negative-class recording. Although extensive signal-domain augmentation was applied, the resulting dataset cannot fully represent the variability of different drone platforms, transmitter hardware, antenna configurations, operating environments, and radio propagation conditions. Consequently, the reported results should be interpreted as validation of the proposed dataset-generation methodology and lightweight CNN architecture rather than proof of universal generalization across all FPV drone systems. Nevertheless, the live GNU Radio evaluation was conducted using a different FPV platform and video transmitter than those used for dataset generation, providing preliminary evidence that the learned features are not limited to a single RF source.
The current binary classification approach assumes a single dominant signal. In the presence of multiple simultaneous transmitters, overlapping patterns may deviate significantly from training data, degrading performance. In real operational scenarios, this is likely to happen, because interference between drones would reduce their controllability. 
Multipath propagation introduces additional variability through reflections, which can alter the rasterized representations compared to the training samples. Changes in transmitter-receiver distance result in a wide range of SNR, which are not fully captured by synthetic noise augmentation. Improving robustness will require more diverse datasets, including varying environments, interference conditions, and signal types, as well as exploration of architectures that better capture temporal and spectral variability.

\subsection*{Computational Trade-offs of Rasterized RF Processing}

Unlike spectrogram-based RF classification approaches, the proposed method eliminates Fourier-based preprocessing entirely. Rasterization consists only of sequentially mapping RF samples into a two-dimensional image representation, yielding linear computational complexity, $\mathcal{O}(n)$, where $n$ denotes the number of processed samples. In contrast, spectrogram generation requires repeated FFT operations with computational complexity $\mathcal{O}(n \log n)$, together with additional windowing and overlap processing.

This comparison represents an order-of-growth analysis rather than an exact runtime estimate, as practical spectrogram implementations depend on FFT window length, overlap ratio, and implementation-specific optimizations. Nevertheless, eliminating the frequency-domain transformation stage simplifies preprocessing prior to classification.

\begin{table}[ht!]
   \caption{Comparison of the proposed rasterization-based CNN approach with representative RF-based drone detection methods reported in the literature. The methods are compared in terms of signal representation, reported detection performance, processing pipeline requirements, and suitability for real-time deployment.}
    \label{tab:4}
    \begin{tabular}{|p{120pt}|p{90pt}|p{70pt}|p{120pt}|p{40pt}|}
        \hline
        \rowcolor{lightgray}
        Method & Representation & Reported Performance & Processing Pipeline & Real-time capability \\
        \hline
        Spectrogram-based CNN \cite{ref32} & Frequency & 96--98\% & STFT + CNN & Medium \\
        \hline
        Multi-channel 1D CNN \cite{ref33} & Feature/ Frequency & 99\% & Feature extraction + CNN & Medium \\
        \hline
        RF-based Detection and Classification \cite{ref34} & Frequency & 97--99\% & STFT + CNN & Low \\
        \hline
        Proposed Rasterized CNN & Time & 99--99.9\% (offline) & Rasterization + CNN & High \\
        \hline
    \end{tabular}
\end{table}

\subsection*{Comparison with Existing Methods}
Frequency-domain approaches, typically based on spectrogram representations, achieve high classification accuracy (96–99\%) under controlled conditions \cite{ref32, ref33}. However, these methods require repeated spectral transformations, increasing computational complexity and latency. In contrast, the proposed rasterization-based approach operates directly in the time domain, eliminating the need for spectral preprocessing.
In terms of model complexity, prior works often employ deep CNN architectures with significantly higher parameter counts, whereas the proposed models achieve comparable accuracy using fewer than 10k parameters. Additionally, the proposed method demonstrates substantially higher inference throughput (up to 3000 images/s), making it suitable for real-time embedded deployment.
However, frequency-domain methods exhibit improved robustness under challenging conditions, particularly in low SNR and multipath environments \cite{ref34}. This highlights a fundamental trade-off: the proposed approach prioritizes computational efficiency and real-time capability, while spectrogram-based methods offer stronger robustness in complex electromagnetic scenarios. Summary of key differences between the proposed approach and representative RF-based drone detection methods reported in the literature can be found in Tab. \ref{tab:4}.

\section*{Conclusion}
This study investigated the applicability of CNN for the automated detection of FPV drone video transmissions in RF streams captured by a SDR-based EW framework. The motivation for this work arose from the limitations of previously implemented signal detection methods, particularly autocorrelation-based approaches, which proved insufficiently robust when operating in dynamic electromagnetic environments. To address these limitations, an alternative detection strategy based on rasterized RF signal representations and CNN classification was explored. A complete workflow was developed that includes RF signal acquisition, rasterization of the signal stream, dataset generation with controlled augmentation, neural network training, and integration into a real-time signal processing environment. Using this workflow, a dataset consisting of approximately 40,000 rasterized RF images was created and used to train multiple custom CNN architectures. The primary objective was to benchmark these architectures to determine how model depth, width, and parameter count influence classification accuracy, computational efficiency, and real-time operational behavior.

The offline benchmarking results demonstrated that nearly all investigated architectures achieved very high classification accuracy, typically exceeding 99\%. However, additional testing within a live GNU Radio signal processing chain revealed that models with similar offline accuracy can behave significantly differently in real-time conditions. Some architectures exhibited excessive sensitivity, producing false detections, while others failed to respond consistently to the presence of the FPV VTX. These findings highlight the importance of validating AI-based detection algorithms within realistic operational environments rather than relying solely on static dataset evaluation. Despite these differences, the results clearly demonstrate that lightweight CNNs can perform reliable RF signal detection. In particular, model C2233 achieved 100\% live-test accuracy despite requiring only 1,077 trainable parameters. An additional advantage of these lightweight models is their compatibility with further optimization techniques. The investigated CNN architectures can be quantized to INT8 precision, significantly reducing memory footprint and computational load without substantially degrading detection performance \cite{ref5}, \cite{ref39}. This makes them well suited for deployment on embedded processors, microcontrollers, or specialized AI accelerators commonly used in field-deployable sensing platforms.

Compared with spectrogram-based RF classification approaches, the proposed method simplifies preprocessing by eliminating Fourier-domain transformations. Although frequency-domain methods may offer greater robustness under challenging propagation conditions, the proposed rasterization-based framework achieves comparable detection performance with substantially lower computational requirements. These characteristics make it particularly attractive for portable and embedded electronic warfare systems, where power consumption, memory footprint, and processing latency are critical design constraints.

Beyond the specific application, the rasterized signal representation combined with CNN-based classification can be applied to other forms of electromagnetic signal detection. One of our research areas focuses on information leakage from electronic equipment, such as unintended RF emissions generated by digital interfaces like HDMI cables driving LCD displays \cite{ref2, ref3}. In particular, preliminary observations show that emissions from LCD display interfaces exhibit raster-like structures comparable to FPV video transmissions, with similar vertical and horizontal patterns appearing in the rasterized representation. This opens the possibility of applying the proposed method for automated detection of information leakage and integrity degradation of TEMPEST equipment, where shielding effectiveness may deteriorate over time due to physical wear or damage.
Another ongoing research effort of ours targets micro-satellite development that includes onboard hyperspectral image processing tasks \cite{ref7, ref9}. In such platforms, strict limitations on power consumption, computational capacity, and communication bandwidth require efficient onboard data processing. The demonstrated balance between computational efficiency and detection capability suggests that tailored CNN models could be used for simple onboard object detection or signal classification tasks. Finally, the observed discrepancy between offline accuracy and real-time behavior highlights a limitation of purely image-based CNN approaches when applied to continuous RF streams. This motivates the investigation of alternative model architectures that explicitly capture temporal dependencies in the signal. Future work will therefore include the evaluation of temporal convolutional networks (TCNs), transformer-based models, and autoencoder-based anomaly detection methods. These approaches may provide improved robustness to temporal variability, interference, and non-stationary signal conditions, enabling more reliable performance in complex electromagnetic environments.


\section*{Funding}

Supported by the DKÖP-NKE-24-06 and DKÖP-NKE-24-08 Doctoral Excellence Scholarship Program of the Ministry for National Research, Development and Innovation Fund.

\section*{Author contributions statement}

G. Farkas and G. Fazekas and P. Karakai performed the dataset generation, CNN model creation, benchmarking and wrote the main manuscript. A. Németh and G. Farkas supervised the research and provided manuscript feedback. All authors reviewed the manuscript.

\section*{Data availability}

The datasets generated and analysed during the current study are available in the Github repository, https://github.com/fg-csp/compact-cnn-drone-detection


\section*{Competing interests} 

The authors declare no competing interests.

\end{document}